\documentclass[conference]{IEEEtran}
\IEEEoverridecommandlockouts
\usepackage{cite}
\usepackage{amsmath,amssymb,amsfonts}
\usepackage{algorithmic}
\usepackage{graphicx}
\usepackage{hyperref}
\usepackage{xcolor}
\usepackage{array}
\usepackage{multirow}
\usepackage{amsmath}
\usepackage{array}
\usepackage{stfloats}
\def\BibTeX{{\rm B\kern-.05em{\sc i\kern-.025em b}\kern-.08em
    T\kern-.1667em\lower.7ex\hbox{E}\kern-.125emX}}
\begin{document}

\title{Contextual Sentence Analysis for the Sentiment Prediction on Financial Data}

\author{\IEEEauthorblockN{Elvys Linhares Pontes}
\IEEEauthorblockA{\textit{Trading Central Labs} \\
\textit{Trading Central}\\
Valbonne, France \\
elvys.linharespontes@tradingcentral.com}
\and
\IEEEauthorblockN{Mohamed Benjannet}
\IEEEauthorblockA{\textit{Trading Central Labs} \\
\textit{Trading Central}\\
Paris, France \\
mohamed.benjannet@tradingcentral.com}
}

\maketitle

\begin{abstract}
Newsletters and social networks can reflect the opinion about the market and specific stocks from the perspective of analysts and the general public on products and/or services provided by a company. Therefore, sentiment analysis of these texts can provide useful information to help investors trade in the market.
In this paper, a hierarchical stack of Transformers model is proposed to identify the sentiment associated with companies and stocks, by predicting a score (of data type real) in a range between --1 and +1. Specifically, we fine-tuned a RoBERTa model to process headlines and microblogs and combined it with additional Transformer layers to process the sentence analysis with sentiment dictionaries to improve the sentiment analysis.
We evaluated it on financial data released by SemEval-2017 task 5 and our proposition outperformed the best systems of SemEval-2017 task 5 and strong baselines. Indeed, the combination of contextual sentence analysis with the financial and general sentiment dictionaries provided useful information to our model and allowed it to generate more reliable sentiment scores.
\end{abstract}


\section{Introduction}
\label{sec:intro}

Sentiment analysis is the study of the emotions expressed by individuals about a product, a service, a brand or more generally a subject on which their attention is focused. 
The challenge of sentiment analysis is to determine in a text the orientation of the opinions expressed about an entity or one of its aspects (a specific element or subdivision of the entity). 
Its goal is to quantify the sentiment about an entity with a negative, neutral or positive value.

In the context of the financial market, the semantic analysis is relevant because previous works showed that sentiments and opinions can affect market dynamics~\cite{Goonatilake2007,VANDEKAUTER20154999}. An entity can represent a stock, a commodity, a foreign exchange market (\textit{Forex}), cryptocurrency or other financial instruments. An article may refer to several entities and the sentiments of these entities may be different from each other. In this case, the sentiment analysis will highlight the corresponding sentiment for each target entity.

The domain of finance has unique linguistic and semantic features. Finance texts are composed of terms which demand precise definitions and many times associated with the quantification of economic phenomena. For instance, the word ``rally" in finance means that large amounts of money are entering the market and pushing stock prices higher; while it can also means a public meeting of protest or support. Moreover, finance texts are composed of newsletters (objective)~\footnote{\url{https://www.reuters.com/}} and microblogs (subjective)~\footnote{\url{https://stocktwits.com/}}~\footnote{\url{https://twitter.com/}} that have different ways to express the information and the opinion about an entity. Therefore, the correct analysis of these texts require common and domain-specific data.

Motivated by the potential use of sentimental analysis to predict the market reaction, the SemEval-2017 task 5 (``\textit{Fine-Grained Sentiment Analysis on Financial Microblogs and News}") proposed the sentiment analysis of financial microblogs and news~\cite{cortis-etal-2017-semeval}. This task is designed to assess general market sentiment as well as sentiment about specific stocks. The best systems proposed hybrid approaches~\cite{jiang-etal-2017-ecnu,ghosal-etal-2017-iitp,mansar-etal-2017-fortia,kar-etal-2017-ritual} by combining deep learning models with sentiment dictionaries to analyse better the sentiment of stocks in financial data. Despite the good results obtained in this task, a more robust model is needed to analyse better financial data and correctly predict their sentiment.

In order to make a deep analysis of finance texts, we propose a system that first generates the contextual word representation of these texts to provide a general analysis of them. Afterwards, we use sentiment dictionaries to provide sentiment scores for each word. Then, a hierarchical stack of Transformers combines these word representations with their corresponding sentiment scores to generate a sentiment representation by considering the sentiment of words in a specific context. Finally, we combine the general and the sentiment representation of a text to generate the sentiment score of an entity in this text. 

The main contributions of this paper are in two-fold. (i) As Transformers models improved the performance of several systems in the sentiment analysis~\cite{hoang-etal-2019-aspect,xu-etal-2019-bert}, we analyse several BERT-based models on the sentiment analysis on financial data.  (ii) We present a new model for sentiment analysis that is based on a hierarchical stack of Transformers to process finance texts and combine this analysis with sentiment dictionaries to generate better sentiment scores for the finance domain. Indeed, our proposition provided more reliable sentiment scores and overcame strong baselines and the best systems of SemEval-2017 task 5.

This paper is organised as follows: we describe related work on sentiment analysis in Section~\ref{sc:rw}. Next, we detail our approach in Section~\ref{sc:architecture}. The experiments and the results are discussed in Sections~\ref{sc:setup} and~\ref{sc:evaluation}. Lastly, we provide the conclusion and some final comments in Section~\ref{sc:conc}.

\section{Related Work}
\label{sc:rw}

Sentiment analysis task aims to analyse a text in order to understand the opinion expressed in it. It can be viewed as a classification or regression task. Most works in sentiment analysis~\cite{rosenthal-etal-2017-semeval} classify the sentiment in three classes: negative, neutral or positive. Other works~\cite{cortis-etal-2017-semeval,mansar-etal-2017-fortia,kar-etal-2017-ritual} aim to precise the sentiment in a float value with range $[-1,+1]$.

Sentiment analysis is one of the most popular applications of natural language processing. It has many real-life applications. To name a few: it can be used to verify the quality of a product by analysing users' reactions to it or the public sentiment in a social network about a topic. For instance, the SemEval-2017 Task 4~\cite{rosenthal-etal-2017-semeval} was focused on the sentiment analysis of tweets on Arabic and English. In this task, participants should classify the sentiment of tweets as negative, neutral or positive, but also the sentiment of tweets toward a topic which can diverges from the sentiment of the overall tweet. 

In order to help the sentiment analysis, several dictionaries were created for general and specific domains. Baccianella et al.\cite{BaccianellaES10} presented the SENTIWORDNET 3.0, a lexical resource explicitly devised for supporting sentiment classification and opinion mining applications. They automatically annotated all WORDNET synsets according to their degrees of positivity, negativity, and neutrality. Bradley and Lang~\cite{Bradley1999AffectiveNF} created the dictionary ``Affective Norms for English Words" (ANEW) in order to provide a set of normative emotional ratings for a large
number of words in the English language. Wilson et al.~\cite{Wilson2005} annotated 15,991 subjective expressions from 425 documents (8,984 sentences) were annotated in positive, negative, both, or neutral. Of these sentences, 28\% contain no subjective expressions, 25\% contain only one, and 47\% contain two or more.

Going to the social network domain, Nielsen~\cite{Nielsen11} annotated a dictionary based on tweets. Its dictionary is composed of 2477 unique words with sentiment scores from $--5$ (very negative) to $+5$ (very positive). 
On the finance domain, Chen et al.~\cite{CHEN18} constructed a market sentiment dictionary based on more than 330K labeled posts crawled from financial social media. Their dictionary contains 8,331 words, 112 hashtags and 115 emojis.

Finance news are composed of a specific vocabulary that may diverges of general meaning. Therefore, general models could not extract the correct meaning for complex finance texts. For instance, Krishnamoorthy~\cite{Krishnamoorthy2017SentimentAO} proposed a sentiment analysis approach that employs the concept of financial and non-financial performance indicators. It presented an association rule mining based hierarchical sentiment classifier model to predict the polarity of financial texts as positive, neutral or negative.

The SemEval-2017 task 5~\cite{cortis-etal-2017-semeval} analysed the fine-grained sentiment analysis on financial microblogs and news. Several participants combined sentiment dictionaries\ (fi\-nance-specific and general domains), as well as custom dictionaries created in the context of this task. Jiang et al.~\cite{jiang-etal-2017-ecnu} and Ghosal et al.~\cite{ghosal-etal-2017-iitp} achieved the best results for the microblog subtask, while Mansar et al.~\cite{mansar-etal-2017-fortia} and Kar et al.~\cite{kar-etal-2017-ritual} for the news headline subtask. 

For the analysis of microblogs, Jiang et al.~\cite{jiang-etal-2017-ecnu} used linguistic, sentiment lexicon, domain-specific and word embedding features, and employ ensemble regression models to predict the sentiment score of a text. Ghosal et al.~\cite{ghosal-etal-2017-iitp} proposed a multilayer perceptron based on an ensemble method that leverages the combination of convolutional and long-short term memory neural networks and feature based models for the sentiment prediction.

For the news headline subtask, Mansar et al.~\cite{mansar-etal-2017-fortia} developed an approach that leverages affective dictionary and word embeddings in combination with convolutional neural networks to infer the sentiment of financial news headlines towards a target company. Kar et al.~\cite{kar-etal-2017-ritual} combined hand-engineered lexical, sentiment, and metadata features with the representation learned from convolutional neural networks and bidirectional gated recurrent unit having attention model applied on top.

Recent works on sentiment analysis have used BERT-based models to improve sentiment analysis~\cite{xu-etal-2019-bert,hoang-etal-2019-aspect,8995193}. Xu et al.~\cite{xu-etal-2019-bert} used the BERT model and proposed a joint post-training approach to enhancing both the domain and task knowledge on the aspect-based sentiment analysis. Hoang et al.~\cite{hoang-etal-2019-aspect} proposed a combined model, which uses only one sentence pair classifier model from BERT to solve both aspect classification and sentiment classification simultaneously.
On the finance domain, Sousa et al.~\cite{8995193} manually annotated 582 stock news from several financial news sources in negative, neutral or negative. Then, they fine-tuned a BERT model on this dataset.

Inspired by the performance of BERT-based models on the sentiment analysis, we propose a method that uses BERT-based model to process finance texts. As the finance domain contains specific vocabulary and sentiment, we enhance our method by providing general and finance dictionaries to improve the analysis of the sentiment expressed by an entity in a text. In order to combine the sentence analysis provided by the BERT-based model and the dictionaries, we stack Transformer encoder layers to process the contextual word representation and the sentiment scores provided by the dictionaries and generate the sentiment score of an entity in the range $[-1,+1]$. 

Similar to our work, Ema et al.~\cite{boros-etal-2020-alleviating} proposed a deep learning model that combines a stack of Transformer layers on top of a fine-tuned BERT encoder and a conditional random fields decoder to recognise named entities on historical newspapers. Our approach is described in the next section.

\section{Our proposition}
\label{sc:architecture}

Our sentiment analysis model is based on pre-trained BERT-based models (Bidirectional Encoder Representations from Transformers)~\cite{devlin-etal-2019-bert}. Although original recommendations suggest that unsupervised pre-training of BERT-based encoders are expected to be sufficiently powerful on modern datasets, we consider that adding extra Transformer layers and sentiment dictionaries could contribute to the analysis of sentence in the finance domain and improve the sentiment analysis of entities. 

\begin{figure}[tb]
    \centering
    \includegraphics[width=\columnwidth]{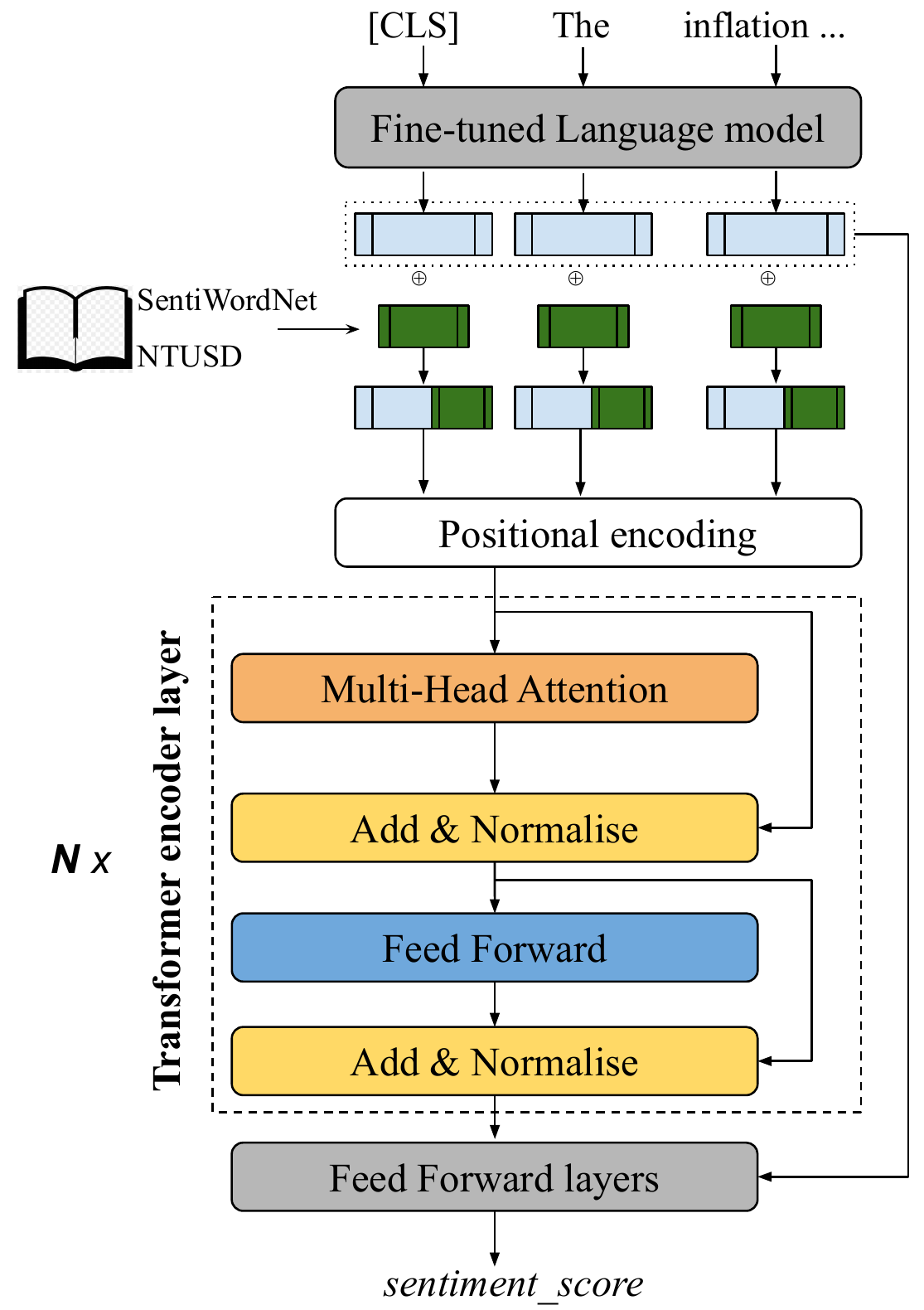}
    \caption{The architecture of our neural network model is composed of: a BERT-based language model, SentiWordNet and NTUSD sentiment dictionaries, a stack of Transformer encoder layers, and a feed-forward layer. The value of sentiment score is included in $[-1,+1]$.}
    \label{fig:architecture}
\end{figure}

First, we use a pre-trained BERT-based model to generate the token representation of a text. After, we use sentiment dictionaries to provide sentiment scores for each token. Then, we stack $n$ Transformer encoder layers on top of this model to process the token representation with their corresponding sentiment scores to generate a sentiment representation. Finally, a feed forward layer processes both general and sentimental representation to define the sentiment score of an entity in the text. 
The global architecture of our model is depicted in Fig. \ref{fig:architecture}. 

The tokenisation process occurs in two steps. First, the built-in tokeniser of BERT-based models performs simple white-space tokenisation, and then WordPiece \cite{wu2016google} splits words into subwords. For example, the word \textit{currencies} can be split into ('cu', \ '\#\#rre', '\#\#ncies'), where \#\# is a special symbol for representing the presence of a sub-word that was recognised. 

In order to provide an additional sentimental information, our model uses sentimental scores from SentiWordNet~\cite{Baccianella2010} and NTUSD~\cite{CHEN18} dictionaries. SentiWordNet is a lexical resource explicitly devised for supporting sentiment classification and opinion mining applications. This resource automatically annotated all WORDNET synsets according to their degrees of positivity, negativity, and neutrality. NTUSD is a market sentiment dictionary based on more than $330k$ labeled posts crawled from financial social media. This dictionary contains $8,331$ words, $112$ hashtags and $115$ emojis. We extract the sentiment scores $Pos$, $Neg$ and $Obj$ from SentiWordNet and the sentiment score from NTUSD for each word in the input. We concatenate the representation generated by BERT-based model for each token (768 dimensions) with their corresponding semantic value to generate a representation of 772 dimensions for each token.

On top of BERT-based model, we add a stack of Transformer encoder layers. A Transformer encoder layer, as proposed in \cite{vaswani2017attention}, is a deep learning architecture based on multi-head attention mechanisms with sinusoidal position embeddings. It is composed of a stack of identical layers. Each layer has two sub-layers. The first is a multi-head self-attention mechanism, and the second is a simple, position-wise fully connected feed-forward network. A residual connection is around each of the two sub-layers, followed by layer normalisation. All sub-layers in the model, as well as the embedding layers, produce outputs of dimension $772$.

We assume that the additional Transformer layers can improve the analysis generated by the BERT-based model by combining this representation with the dictionaries and contribute to the semantic prediction of sentences. Finally, a feed-forward layer concatenates the contextual word representation of the [CLS] token from the stack of Transformer blocks and the [CLS] token from BERT-based model to generate the sentiment score with range $[-1, 1]$.

\section{Experimental setup}
\label{sc:setup}

\subsection{Datasets}

Following the same idea of SemEval-2017 task 5, we conduct the experiments on the official
datasets constructed by SSIX project\footnote{http://ssix-project.eu/}~\cite{davis2016}. These datasets are composed of two types of news: microblogs and headlines.

The microblog dataset consists of a collection of financially relevant microblog messages from Twitter\footnote{\url{https://twitter.com}} and StockTwits\footnote{\url{https://stocktwits.com/}}. StockTwits is a social media platform designed for sharing ideas between investors, traders, and entrepreneurs. Its messages focus on stock market events and typically contain references to company stock symbols. The news headlines dataset is composed of news headlines extracted from different sources on the Internet, such as AP News, Reuters, Forbes and Handelsblatt. Some examples of these datasets are shown in Table~\ref{tab:example}. The entity column represents the company/stock analysed on the text column. 

Both datasets were manually annotated by using a floating-point value between $-1$ (very bearish/negative) and $+1$ (very bullish/positive). Messages containing information that reveals a positive trend for a company or stock are annotated with positive values, while messages implying a negative trend are annotated with negative sentiment scores. 

Table~\ref{tab:dataset} summarises the statistics about these datasets. Microblogs are composed of short texts (a few words), while the news headlines contain longer and more structured texts. Both datasets contain unbalanced data, i.e. the number of negative and positive examples are not equivalent. Positive examples are 50\% more numerous than negative ones for the news headline dataset. This difference is even greater for the microblogs dataset (about 98\%).

\begin{table*}[tb]
\caption{Examples of microblogs and news headlines extracted from SSIX dataset. }
\label{tab:example}       
\begin{center}
\begin{tabular}{|l|l|l|p{8cm}|c|}
\hline
\textbf{Source}             & \multicolumn{1}{c}{\textbf{}} & \multicolumn{1}{c}{\textbf{Entity}} & \multicolumn{1}{c}{\textbf{Text}} & \textbf{Sentiment} \\
\hline
\multirow{4}{*}{Microblogs} & Twitter & \$DIA & Lunchtime rally coming & 0.46 \\\cline{2-5}
                            & stocktwits & \$SPY & due to slow growth & -0.159 \\\cline{2-5}
                            & stocktwits & \$RTN & Moved Upper Bollinger Band & 0.498 \\\cline{2-5}
                            & stocktwits & \$RDEN & Very tight price & 0 \\\hline
\multicolumn{2}{|l|}{\multirow{3}{*}{News headlines}}         & Morrisons & Morrisons book second consecutive quarter of sales growth & 0.43 \\\cline{3-5}
\multicolumn{2}{|l|}{} & Barclays & Barclays 'bad bank' chief to step down & -0.231 \\\cline{3-5}
\multicolumn{2}{|l|}{} & Schroders & Schroders posts FY profit beat, replaces CEO and chairman in board shake-up & 0.052 \\
\hline
\end{tabular}
\end{center}
\end{table*}

\begin{table}[tb]
\caption{Statistics of training and test datasets of two subtasks. The third and fourth columns represent the total of examples and the average number of words by example, respectively. The symbols $\$$ represents the sentiment scores. The last columns represent the amount of examples with the sentiment score smaller and bigger (or equal) than zero, respectively.}
\label{tab:dataset}       
\begin{center}
\begin{tabular}{|l|c|c|c|c|c|}
\hline
\textbf{Type} & \textbf{Split} & \textbf{\#Total} & \textbf{Length} & \textbf{$\$<0$} & \textbf{$\$>=0$}  \\ \hline
\multirow{2}{*}{StockTwits} & train & 934 & 6.4 & 333 & 601 \\ \cline{2-6}
          & test & 429 & 6.5 & 141 & 288 \\\hline
\multirow{2}{*}{Twitter}    & train & 766 & 5.7 & 248 & 518\\ \cline{2-6}
          & test & 365 & 5.4 & 116 & 249\\\hline
\multirow{2}{1.4cm}{News headline} & train & 1142 & 9.6 & 451 & 691\\ \cline{2-6}
         & test & 491 & 9.5 & 203 & 288  \\
\hline
\end{tabular}
\end{center}
\end{table}

\subsection{Evaluation measures}

Following the same evaluation procedure of the SemEval-2017 task 5, the systems were evaluated based on the cosine similarity that compares the proximity between gold standard and predicted results. Cosine similarity is calculated according to equation (\ref{eq:cosine}), where $G$ is the vector of gold standard scores and $P$ is the vector of corresponding scores predicted by the system:

\begin{equation}
    \label{eq:cosine}
    cosine(G,P) = \frac{\sum\limits_{i=1}^{n} G_i \times P_i}{\sqrt{\sum\limits_{i=1}^{n}G_i^2} \times \sqrt{\sum\limits_{i=1}^{n} P_i^2}}
\end{equation}

The final score (\ref{eq:final}) is obtained by weighting the cosine similarity with the ratio of scored instances to reward systems that attempt to answer all problems in the gold standard~\cite{ghosh-etal-2015-semeval}.

\begin{equation}
    \label{eq:final}
    score(G,P) = \frac{|P|}{|G|} \times cosine(G, P)    
\end{equation}

\subsection{Baselines}

\begin{figure}[tb]
    \centering
    \includegraphics[width=0.7\columnwidth]{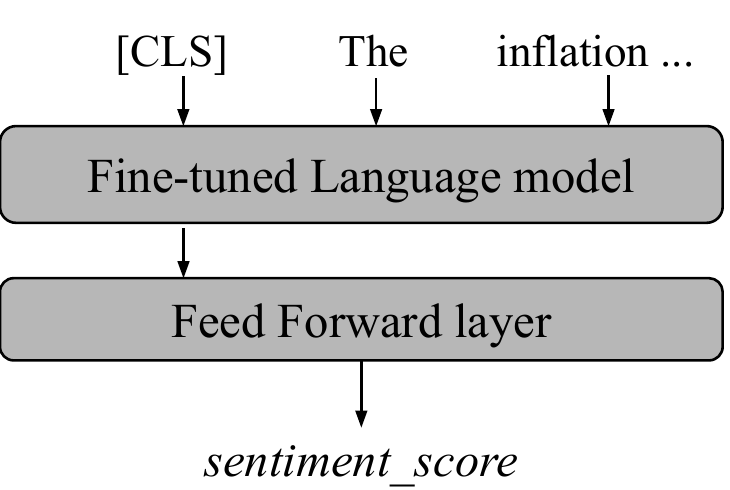}
    \caption{Architecture of our BERT-based baselines. The feed-forward layer receives the [CLS] token representation and generate an output with a node (sentiment score).}
    \label{fig:baseline}
\end{figure}

We analysed as baselines the three most used BERT-based language models: BERT~\cite{devlin-etal-2019-bert}, RoBERTa (Robustly optimized BERT approach)~\cite{liu2019roberta}, and DistilBERT (a distilled version of BERT)~\cite{sanh2020distilbert}. We took the representation of the [CLS] token at the last layer of these models and we added a feed forward layer to generate a sentiment score for a input text (Fig.~\ref{fig:baseline}). 

In order to compare the performance of our approach with the state of the art, we also selected the best three systems for the microblogs and news headline datasets at SemEval-2017 task 5.

\subsection{Parameters}

In order to choose the best BERT-based model for our architecture, we first trained the baselines and we selected the best model (Tables~\ref{tab:results:microblogs} and~\ref{tab:results:headlines}). In this case, RoBERTa baseline outperformed the other baselines by achieving better cosine scores. 

Several tests were done with different amount of Transformer encoder layers (range from 1 to 6) to generate different levels of the sentiment representation. We refer to our model as RoBERTa$+n\times$Transf, where $n$ is the number of Transformer encoder layers. All models were trained by using batch normalisation, mean squared error loss and Adam optimiser. The learning rate and the Adam epsilon were set to $10^{-5}$ and $10^{-6}$, respectively. All systems were trained for each kind of data (news headlines and microblogs).

For the preprocessing, all HTTP links were removed from the texts of microblogs and news headlines for the training procedure.

\section{Experimental evaluation}
\label{sc:evaluation}

In this section, we provide the experimental results of the baseline models and the proposed method. Tables~\ref{tab:results:microblogs} and~\ref{tab:results:headlines} show the performance of all models for the microblog and news headline datasets, respectively.
We compare the best systems of SemEval-17 task 5 (first part of tables), baselines (second part of tables) and all versions of our proposed model (last part of tables).

\begin{table}[tb]
\caption{Cosine similarity scores for the microblog dataset. The results of baselines and all versions of our approach were averaged over 5 runs. Best results are in bold.}
\label{tab:results:microblogs}       
\begin{center}
\begin{tabular}{|l|c|}
\hline
\textbf{Models} & \textbf{Cosine} \\
\hline
Jiang et al.~\cite{jiang-etal-2017-ecnu} & 0.778 \\
Ghosal et al.~\cite{ghosal-etal-2017-iitp}    & 0.751 \\
Deborah et al.~\cite{s-etal-2017-ssn-mlrg1}  & 0.735 \\
\hline
BERT        & 0.789  \\
DistilBERT  & 0.784  \\
RoBERTa     & 0.832  \\
\hline
RoBERTa$+1\times$Transf & 0.833 \\
RoBERTa$+2\times$Transf & 0.832 \\
RoBERTa$+3\times$Transf & 0.831 \\
RoBERTa$+4\times$Transf & 0.834 \\
RoBERTa$+5\times$Transf & \textbf{0.841} \\
RoBERTa$+6\times$Transf & 0.833 \\
\hline
\end{tabular}
\end{center}
\end{table}

All BERT-based models outperformed the best systems of SemEval-2017 which shows the potential of BERT-based models in analysing sentiment on finance texts. As expected, BERT achieved better results than DistilBERT. Indeed, DistilBERT has less parameters (around 40\% less) than BERT which explains the limitation in the sentiment analysis. RoBERTa achieved better performance than BERT. In fact, the pre-trained RoBERTa model is built on the BERT model with different key hyper-parameters, without the next-sentence pre-training objective and trained with much larger mini-batches and learning rates, which allowed it to perform better than BERT on several natural language processing tasks~\cite{liu2019roberta}.

\begin{table}[tb]
\caption{Cosine similarity scores for the news statements \& headlines dataset. The results of baselines and all versions of our approach were averaged over 5 runs. Best results are in bold.}
\begin{center}
\label{tab:results:headlines}       
\begin{tabular}{|l|c|}
\hline
\textbf{Models} & \textbf{Cosine} \\
\hline
Mansar et al.~\cite{mansar-etal-2017-fortia} & 0.745 \\
Kar et al.~\cite{kar-etal-2017-ritual}    & 0.744 \\
Rotim et al.~\cite{rotim-etal-2017-takelab}  & 0.733 \\
\hline
BERT        & 0.790  \\
DistilBERT  & 0.782  \\
RoBERTa     & 0.839  \\
\hline
RoBERTa$+1\times$Transf & 0.844 \\
RoBERTa$+2\times$Transf & 0.847 \\
RoBERTa$+3\times$Transf & \textbf{0.848} \\
RoBERTa$+4\times$Transf & 0.842 \\
RoBERTa$+5\times$Transf & \textbf{0.848} \\
RoBERTa$+6\times$Transf & 0.843 \\
\hline
\end{tabular}
\end{center}
\end{table}

The generation of the sentiment representation provided additional features to our model and enabled it to achieve the best cosine results for almost all sentiment representation levels. Indeed, adding Transformer encoder layers with sentiment dictionaries allowed our model to perform better on both metrics. For both subtasks, the addition of transformers kept or improved the performance compared to the RoBERTa baseline with exception of RoBERTa $+3\times$Transf for the microblog subtask. 

As shown on Tables~\ref{tab:results:microblogs} and~\ref{tab:results:headlines}, the stack of $5$ Transformer encoder layers generated the best sentiment representation and allowed our model to achieve the best performance for both subtasks. The version RoBERTa $+3\times$Transf also achieved the best results for the news headline subtask; however, the RoBERTa baseline outperformed this version for both metrics on the microblog dataset. 

\begin{table*}[t]
\centering
\caption{Study case of baselines and the best version of our system. GS means the gold standard.}
\label{tab:pred_examples}       
\begin{tabular}{|l|p{1.4cm}|p{5cm}|c|c|c|c|m{1.4cm}|}
\hline
\textbf{Source} & \textbf{Entity} & \textbf{Text} & \textbf{GS} & \textbf{BERT} & \textbf{DistilBERT} & \textbf{RoBERTa} & \textbf{RoBERTa $+5\times$Tranfs} \\
\hline
stocktwits & \$UNG & Bot more \$UNG puts & -0.875 & -0.027 & -0.139 & -0.429 & \textbf{-0.497} \\\hline
stocktwits & \$C & As Alcoa Kicks Off March Quarter Earnings, We See More Reasons To Be Cautious https://t.co/QIPEseBq6s \$FB \$AA \$AAPL \$BAC \$BKE \$BLK \$C \$CAT & -0.093 & \textbf{-0.082} & -0.060 & -0.042 & -0.074 \\\hline
stocktwits    & \$GDX & \$GDX \$GDXJ \$JNUG - strong move today for the Junior Gold Miners - keep an EYE out for a gap fill - & 0.750 & 0.707 & 0.659 & 0.698 & \textbf{0.709} \\\hline
\multirow{3}{1.4cm}{News headline} & Barclays & Barclays faces another heavy forex fine & -0.834 & -0.604 & -0.379 & -0.652 & \textbf{-0.672} \\\cline{2-8}
       & Royal Mail & CompaniesRoyal Mail adds a penny to stamp prices &  -0.05 & 0.198 & 0.132 & \textbf{-0.018} & -0.003 \\\cline{2-8}
       & Ashtead & Ashtead to buy back shares, full-year profit beats estimates & 0.588 & 0.453 & 0.300 & 0.724 & \textbf{0.639} \\
\hline
\end{tabular}
\end{table*}

Another interesting point is the complexity of the text. News statements \& headlines dataset is composed of well-structured sentences, while microblogs dataset contains noisy texts (incomplete and informal sentences) that contain several URL links and symbols (Tables~\ref{tab:example} and~\ref{tab:pred_examples}). Despite this noisy text, our model achieved almost the same results for both datasets, showing the ability of our model to adapt to different data types.

\subsection{Study case}

In order to extend the analysis of the systems, we listed some negative, neutral and positive examples of microblogs and news headlines with the gold standard and predicted score for each model (Table~\ref{tab:pred_examples}). 

As you can see, the texts of microblogs are not grammatically correct and contain links and several entities (cashtags), while the texts of news headlines contain correct sentences. These differences in their data require a specific model for each type of text.

Our approach predicted almost all scores closer to the gold standard (GS in the Table) than the baselines. Our method did not get the best predictions for neutral scores but our predictions still closer to the gold standard (less than 0.05 on cosine measure). As expected, the predictions of RoBERTa and our method are similar but our results are slightly better.

The addition of Transformer encoder layers and the dictionaries enabled our model to analyse better the texts and, consequently, generate better predictions. Although our architecture has surpassed the state of the art, there is still a gap to be filled to improve the models. For instance, the difference between our prediction and the gold standard for the negative stocktwits and news headline examples were 0.378 and 0.162 in absolute values, respectively. 

A possible reason for this gap in negative examples is related to the unbalanced data (Table~\ref{tab:dataset}), i.e. the number of negative examples is not sufficient to train our model to generate accurate predictions. Table~\ref{tab:results:values} shows the cosine scores for the positive and negative predictions of the version RoBERTa$+5\times$Transf. For both datasets, our version got better results for the positive predictions than negative as described in Table~\ref{tab:pred_examples}. 

The difference between positive and negative cosine results ($0.02$ and $0.038$ for news statements and microblogs, respectively) is smaller for the news statements \& headlines which has a more balanced training data. Indeed, the number of positive examples is $1.5$ times bigger than the negative for the news statements \& headlines, while this difference is $1.9$ for the microblogs dataset. 

\begin{table}[tb]
\caption{Cosine similarity scores of the version RoBERTa$+5\times$Transf grouped by positive and negative prediction values.}
\begin{center}
\label{tab:results:values}       
\begin{tabular}{|p{3cm}|c|}
\hline
\textbf{Prediction values} & \textbf{Cosine} \\
\hline
\multicolumn{2}{|l|}{\textit{Microblogs}} \\ \hline
positive ($>=0$)  & 0.853 \\
negative  & 0.815 \\
\hline
\multicolumn{2}{|l|}{\textit{News statements \& headlines}} \\\hline
positive ($>=0$) & 0.860 \\
negative  & 0.840 \\
\hline
\end{tabular}
\end{center}
\end{table}

\section{Conclusion}
\label{sc:conc}

Sentiment analysis in the finance domain aims to analyse the sentiment of news related to an entity (company/stock) by classifying them in range $[-1,+1]$, where $-1$ is very negative and $+1$ is very positive. BERT-based models proved to be interesting for dealing with the sentiment analysis of texts~\cite{xu-etal-2019-bert,hoang-etal-2019-aspect,8995193}. Indeed, all BERT-based baselines achieved better results than the best systems of SemEval-2017 task 5 on finance texts.

We presented a deep learning architecture for sentiment analysis based on stacked Transformer layers that includes a fine-tuned RoBERTa encoder and several Transformer encoder blocks. 
The combination of the general representation with the sentiment representation generated by the stack of Transformer encoder layers with sentiment dictionaries provided more accurate sentiment scores for an entity in a text.

Among all versions of our proposition, the version RoBERTa$+5\times$Transf generated the best sentiment representation and achieved the best cosine scores for both subtasks. This model showed its robustness by obtaining the best results for well-structured and noisy texts.
It outperformed the best systems of SemEval-2017 task 5 (13.8\% for new headlines and 8\% for microblogs) and all baselines.

Further works will be done in how to analyse the sentiment of companies in long news articles with complex sentence structures and containing several companies with different sentiment. Another objective can be to annotate more negative examples to balance SSIX datasets. Finally, we also would like to work on the sentiment analysis of stocks in order to predict the evolution of stock prices based on their sentiment.


\bibliographystyle{spmpsci}      
\bibliography{references}   

\end{document}